\documentclass[runningheads]{llncs}
\usepackage[T1]{fontenc}
\usepackage{graphicx}
\usepackage{times}
\usepackage{latexsym}
\graphicspath{ {./Figures/} }
\usepackage{amssymb}
\usepackage{multirow}
\usepackage{rotating}
\usepackage[export]{adjustbox}
\usepackage{tikz}
\usepackage{float}
\usepackage{xcolor,colortbl}
\begin{document}
\title{Assessing Logical Reasoning Capabilities \\ of Encoder-Only Transformer Models}
%
%
\author{Paulo Pirozelli\inst{1}\orcidID{0000-0002-4714-287X} \and Marcos M. José\inst{3}\orcidID{0000-0003-4663-4386}  \and \\ Paulo de Tarso P. Filho\inst{2}\orcidID{0009-0000-8085-795X} \and Anarosa A. F. Brandão\inst{2}\orcidID{0000-0001-8992-4768} \and Fabio G. Cozman\inst{2}\orcidID{0000-0003-4077-4935}}

\authorrunning{Pirozelli et al.}
%
\institute{
Instituto Mauá de Tecnologia \and
Universidade de São Paulo \and
University of Alberta \\
\email{paulo.silva@maua.br}}
\maketitle              
\begin{abstract}
Transformer models have shown impressive  abilities in natural language tasks such as text generation and question answering. Still, it is not clear whether these models can successfully conduct a rule-guided task such as logical reasoning. In this paper, we investigate the extent to which encoder-only transformer language models (LMs) can reason according to logical rules. We ask whether these LMs can deduce theorems in propositional calculus and first-order logic, if their relative success in these problems reflects general logical capabilities, and which layers contribute the most to the task. First, we show for several encoder-only LMs that they can be trained, to a reasonable degree, to determine logical validity on various datasets. Next, by cross-probing fine-tuned models on these datasets, we show that LMs have difficulty in transferring their putative logical reasoning ability, which suggests that they may have learned dataset-specific features instead of a general capability. Finally, we conduct a layerwise probing experiment, which shows that the hypothesis classification task is mostly solved through higher layers.

\end{abstract}
\section{Introduction}


Transformer models are remarkably effective at a wide range of natural language processing (NLP) tasks, such as question answering, summarization, and text generation. By and large, these abilities are the result of specific training processes, where a language model is fine-tuned on a task-specific dataset.
Curiously, encoder-only transformer models (LMs) also exhibit \textit{implicit} linguistic and cognitive abilities for which they were not directly supervised. 
Such LMs have been shown to encode information on tense and number \cite{conneau-etal-2018-cram}, anaphora and determiner-noun agreement \cite{warstadt-etal-2020-blimp-benchmark}, semantic roles \cite{tenney2019whatdoyoulearn}, syntactic dependencies \cite{hewitt2020emergent}, relational knowledge \cite{dai2021knowledgeneurons}, and spatiotemporal representation \cite{gurnee2023language}.



Given that logical reasoning is a core component of intelligence, human or otherwise \cite{norvig2002modern}, it is worth investigating the capabilities of encoder-only transformer models in executing tasks that necessitate such reasoning. Understanding whether LMs can solve logical problems, and the manner in which they tackle these problems, may enable us to understand the extent to which their inferences arise from reasoning rather than purely associative memory. This understanding is crucial for developing mechanisms that facilitate the generation of consistent outputs, whether in a neural-symbolic fashion or through the improvement of model architectures.

The goal of this paper is, thus, to assess the ability of encoder-only transformer models to reason according to the rules of logic --- understood here as deductive arguments expressable in propositional calculus or first-order logic. We examine three main questions throughout this paper: i) Can enconder-only transformer models perform logical reasoning tasks?; 
ii) How general is this ability?; and
iii) What layers better contribute to solving these tasks? 

Section \ref{section:Background} reviews the work on transformers' logical reasoning abilities, as well as the function of probing in uncovering latent knowledge. Next, we gather and describe four datasets grounded on logical deduction (sec. \ref{sec:datasets}). In a first batch of experiments, we conduct a systematic comparison of encoder-only transformer models on these datasets (sec. \ref{sec:hypothesis}). Section \ref{sec:probing} then investigates whether the performance in the previous task could be attributed to some general ability and whether the reasoning learned from one dataset could be transferred to a similar dataset. Finally, we perform a layerwise probing to understand which layers are responsible for solving logical deduction problems (sec. \ref{section:locus}).\footnote{Data, code, and complete results are available at \url{github.com/paulopirozelli/logicalreasoning}.} 

\section{Related Work}\label{section:Background}


\paragraph{Logical Reasoning in Transformer Models}\label{sec:related_logic}
Transformer models are powerful enough to solve logical reasoning tasks expressed in natural language \cite{RuleTaker,SimpleLogic,LogicNLI,FOLIO}. Yet, it is not clear if these models have actually mastered logical reasoning. LMs seem to inevitably rely on statistical artifacts to deduce theorems, rather than on general, rule-based relationships \cite{SimpleLogic}. They  use shortcuts to solve hypothesis classification problems,   leading to vulnerabilities in reasoning (e.g., LMs are fooled when hypotheses appear within rules), and making them susceptible to irrelevant (logically consistent) perturbations \cite{gaskell2022logically}. 

Studies focused on functional words close to logical operators have identified similar shortcomings in LMs' reasoning capabilities.
Transformers struggle to deal with negation, predicting similar probabilities to a sentence and its negation \cite{kassner-schutze-2020-negated,hossain-etal-2020-analysis}.
\cite{kalouli-etal-2022-negation} extend these findings to conjunction, disjunction, and existential and universal quantification, showing that expressions associated with these operations are frequently dominated by semantically rich words.
Transformers also fail in modeling semi-functional-semi-content words in general, such as quantifiers and modals \cite{sevastjanova-etal-2021-explaining}.
Tangentially to logic, \cite{dankers-etal-2022-paradox} show that transformer models do not always properly handle    compositionality: sometimes a translation is more local than desired (treating idioms as regular constructions), sometimes it is excessively global (paying attention to irrelevant parts of the sentence). 
Moreover, there is evidence that the way LMs compose sentences does not align well with human judgment \cite{liu2022representations}.

Large LMs have also been assessed for their logical reasoning capabilities. Despite their impressive achievements in numerous tasks, these models still struggle with multi-step and compositional problems \cite{gopher,dziri2023faith}. Although good at individual deductions, large LMs struggle with proof planing: when many valid proof steps are available, they often take the wrong path \cite{saparov2023language}.
Large LMs also appear to suffer from human-like biases in logical tasks: 
they perform significantly worse when the semantic content is too abstract or conflicts with prior knowledge \cite{dasgupta2022reasoning,tang2023large}.





\paragraph{Probing Tasks}\label{sec:related_probing}

Probing is a technique used to discover if a model has acquired certain type of knowledge. In probing, a dataset that encodes a particular property (e.g., part-of-speech) is used to train a classifier (the probe), taking the representations produced by the original model as inputs for the classification \cite{belinkov-2022-probing}. As the LM is not further trained on the task, as it would be in fine-tuning, the probe's performance depends on whether the information about that property had already been encoded by the model. Thus, success in the task provides some evidence that the model has stored such knowledge in its parameters. Probing can be used to examine several components of LMs, such as embeddings, attention heads, and feedforward computations.

Transformer models have been extensively studied through probing \cite{belinkov-2022-probing}. Most attention has been given to BERT, which gave rise to a
large literature on the properties encoded by this model \cite{rogers-etal-2020-primer}.
RoBERTa has also been studied in some detail through probing; e.g., what abilities that model learns during training \cite{zhang-etal-2021-need,liu2018probing} and its knowledge of semantic structural information \cite{Wu_2021}.



\begin{figure}[h]
 \centering
 \includegraphics[width=0.6\textwidth]{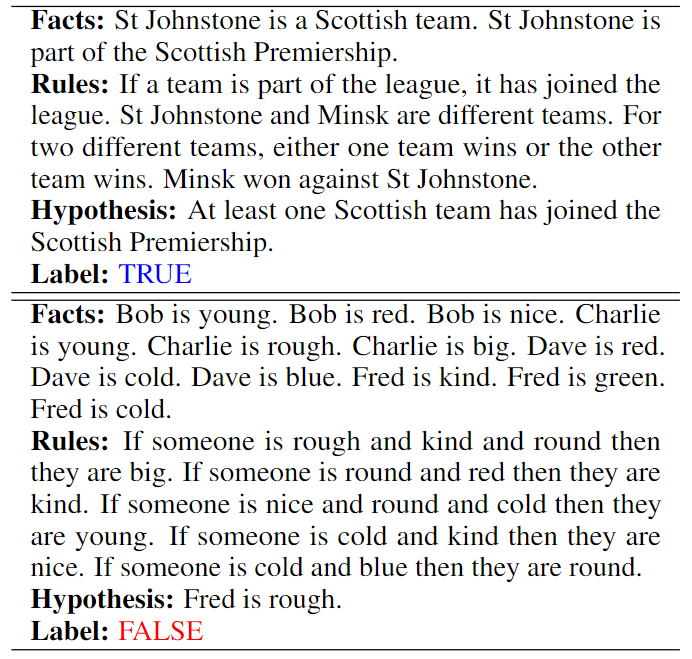}
 \caption{Examples of logical reasoning arguments. The argument at the top is from FOLIO, a manually-written dataset; the one at the bottom is from RuleTaker, a dataset that uses a semi-synthetic approach.}
 \label{fig:examples}
\end{figure}

\section{Logical Reasoning Datasets}\label{sec:datasets}

\begin{table*}[t]
\caption{Main features of the logical reasoning datasets. FOL stands for first-order logic, PC for propositional calculus, and CI for conjunctive implication. The labels in the datasets are as follows: \textbf{FOLIO} (False, True, Unknown), \textbf{LogicNLI} (Contradiction, Entailment, Neutral, Paradox), \textbf{RuleTaker} (False, True), \textbf{SimpleLogic} (False, True). The average number of premises and the average number of words per argument refer to the training set statistics. Appendix \ref{appendix:label} shows the full dataset label distribution.}
\label{tab:datasets}
\vspace*{-1ex}
\centering
\begin{tabular}{l|l|l|l|l|l|l}
\hline
\textbf{Dataset} & \textbf{Size (train/val/test)} & \textbf{Scope} & \textbf{Type} & \textbf{Label} & \textbf{Avg. Premises} & \textbf{Avg. Words}\\ \hline
\textbf{FOLIO} & 1003 / 204 / - & FOL & Manual & 3 & 5.23 & 64.01\\
\textbf{LogicNLI} & 16000 / 2000 / 2000 & FOL & Semi-synt. & 4 & 24 & 211.86\\
\textbf{RuleTaker} & 27363 / 3899 / 7793 & FOL (CI) & Semi-synt. & 2 & 16.30 & 100.06\\
\textbf{SimpleLogic} & 11341 / 1418 / 1417 & PC (CI) & Synthetic & 2 & 60.76 & 467.45\\
\hline
\end{tabular}
\end{table*}

In assessing logical reasoning, we restricted our analysis to datasets related to propositional calculus (PC) and first-order logic (FOL), with FOL being an extension of PC that includes predicates and quantifiers. These logical systems offer a powerful formalism that balances simplicity and expressivity in representing and reasoning about statements and relationships. They are suitable for capturing a wide range of knowledge and formalizing a large number of domains.

In addition to being expressible in PC or FOL, we selected datasets that satisfied three other properties: i) observations had to be as self-contained as possible; ii) sentences needed to have corresponding translations in both logical formalism and natural language; and iii) hypotheses had to be declarative sentences. The first property aims to decouple logic from background knowledge in order to assess pure reasoning capabilities. While using natural language examples means implicit knowledge can never be completely erased, we opted for datasets that minimized this by explicitly stating prior knowledge or by using inference patterns where resorting to prior knowledge is unnecessary. For this reason, we only considered pure logical datasets and did not include other forms of reasoning such as scientific reasoning \cite{clark2018think,RACE,ReClor}, mathematical reasoning \cite{GSM8K,SVAMP,hendrycks2021measuring}, counterfactual reasoning \cite{TimeTravel,yang2020semeval,o2021wish}, planning \cite{valmeekam2023large}, inductive reasoning \cite{sinha-etal-2019-clutrr,tang2023large}, and abductive reasoning \cite{tang2023large}. The second property, the translation into logical formalism, allows one to determine unambiguously the logical relationship between premises and hypotheses, while the natural language counterpart allows us to probe the LMs. Hence, we excluded datasets that lacked natural language translations, such as LTL \cite{LTL2022}. The last property excluded QA datasets \cite{LogiQA,talmor-etal-2019-commonsenseqa}, as they required the understanding of several types of questions (e.g., who, where) that are entangled with semantic and contextual knowledge (e.g., that ``Alice'' is the name of a person).


In the end, four logical reasoning datasets were selected: FOLIO \cite{FOLIO}, LogicNLI \cite{LogicNLI}, RuleTaker \cite{RuleTaker}, and SimpleLogic \cite{SimpleLogic}. Examples of arguments can be seen in Figure \ref{fig:examples}. These datasets cover a wide range of variations in terms of construction (manual, semi-synthetic, synthetic), alignment with common sense, linguistic variability, and scope (PC, full or partial FOL). For instance, FOLIO is human-written; LogicNLI and RuleTaker both use a template which is then manually edited; and SimpleLogic is fully synthetic. FOLIO and LogicNLI encompass the full spectrum of FOL; RuleTaker is expressible in FOL but only covers negation and conjunctive implications,\footnote{Conjunctive implications are arguments of the form ($fact. [\land fact.]^* rule. [\land rule.]^* \Rightarrow hypothesis$).} and SimpleLogic is restricted to conjunctive implications in PC. Regarding the number of labels, RuleTaker and SimpleLogic have True and False labels; FOLIO includes an Unknown label; and LogicNLI admits a fourth possibility, Paradox, where both a hypothesis and its negation can be deduced from the premises. Table \ref{tab:datasets} summarizes the main statistics of the datasets. Appendix \ref{appendix:datasets} provides a detailed description of the four datasets.

For all datasets, inputs are formatted as ``{\tt fact$_1$. fact$_2$. ... fact$_n$. rule$_1$. rule$_2$. ... rule$_m$ <SEP> hypothesis.}'', for which a label must be predicted. The output is simply a probability distribution over the possible labels, which varies from dataset to dataset.







\section{Testing LMs for Hypothesis Classification}\label{sec:hypothesis}



The performance of LMs in logical reasoning, including in the datasets described in the previous section, has been studied in scattered experiments without a clear unified context that allows direct comparison. Due to this, we decided to fine-tune a wide range of pretrained encoder-only transformer models on those datasets.\footnote{We opted to explore encoder-only models because this type of architecture is well-suited for classification tasks. These models have access to the whole input sequence and are typically trained on discriminative tasks, such as masked language modeling.} We modeled this problem as a \textit{hypothesis classification task}, where the goal is to determine the logical relationship between a set of premises and a conclusion. To ensure comprehensive coverage, eight families of LMs were assessed: DistilBERT \cite{sanh2020distilbert}, BERT \cite{BERT2018}, RoBERTa \cite{liu2019roberta}, Longformer \cite{longformer}, DeBERTa \cite{deberta}, AlBERT \cite{lan2019albert}, XLM-RoBERTa \cite{conneau2020unsupervised}, and XLNet \cite{yang2020xlnet}. The full list of models is displayed in the left column of Table~\ref{tab:finetuning}.


%


To fine-tune our models, we used a classification head with a single linear layer of dimension (embedding-length, labels), and applied a dropout of 0.5 to the linear operation. As input for the classification head, we used the last hidden embedding of the [CLS] token, as is customary for classification tasks in NLP. Sequences were padded and truncated to the maximum length allowed by each LM. Models were trained for up to 50 epochs, using early stopping with a patience of 5 epochs. We used Adam as our optimizer (with $\beta$ = [0.9, 0.999] and no weight decay) and experimented with two different learning rates (1e-5 and 1e-6). Models were selected based on the loss for the validation set, with results reported for the test set. The only exception is FOLIO, where only the validation set is publicly available; results are thus reported for this set. We report accuracy as the standard metric, as classes in the datasets are balanced. Table \ref{tab:finetuning} displays the best result achieved in each case. The best and worst scores for each dataset are highlighted in blue and red, respectively. We also provide a largest class baseline for comparison.

\begin{table}[h]
 \caption{Accuracy comparison among several encoder-only transformer models for the hypothesis classification task across four datasets: FOLIO, LogicNLI, RuleTaker, and SimpleLogic. The models are listed in the left column of the table. The best and worst scores for each dataset are highlighted in blue and red, respectively. ``Largest class'' refers to the accuracy achieved by always selecting the class with the highest frequency in the training set. Results for RoBERTa-large, selected for subsequent analyses, are in bold.}
 \label{tab:finetuning}
 \vspace*{-1ex}
 \centering
 \begin{tabular}{l|c|c|c|c}
 \cline{2-5}
 & \textbf{  FOLIO       } & \textbf{  LogicNLI  } & \textbf{  RuleTaker  } & \textbf{SimpleLogic}
 \\
 \hline 
 \textbf{DistilBERT} & 50.98 & 36.20 & 86.55 & 93.58\\
 \hline
 \textbf{BERT base} & 52.45 & 48.75 & 98.91 & 92.31\\
 \textbf{BERT large} & 61.76 & 42.50 & 99.94 & 91.88\\
 \hline
 \textbf{RoBERTa base} & 58.82 & 62.90 & 98.04 & 92.87\\
 \textbf{RoBERTa large} & \textbf{64.71} & \textbf{72.70} & \textbf{99.78} & \textbf{90.83}\\
 \hline
 \textbf{Longformer base} & 62.75 & 58.05 & 99.92 & 94.21 \\
 \textbf{Longformer large} & 62.25 & 74.15 & 99.94 & 92.37 \\
 \hline
 \textbf{DeBERTa xsmall} & 54.41 & 65.30 & 99.81 & 91.67 \\
 \textbf{DeBERTa small} & 53.43 & 59.65 & 95.23 & 93.23 \\
 \textbf{DeBERTa base} & 60.78 & 66.70 & 99.81 & 93.72\\
 \textbf{DeBERTa large} & 64.71 & 84.70 & 99.97 & 93.72 \\
 \textbf{DeBERTa xlarge} & 62.25 & \textcolor{red}{25.00} & \textcolor{blue}{99.99} & \textcolor{red}{49.96} \\
 \textbf{DeBERTa xxlarge} & \textcolor{blue}{71.57} & \textcolor{red}{25.00} & \textcolor{red}{50.02} & 50.04 \\
 \hline
 \textbf{ALBERT base} & 57.35 & 66.80 & 99.91 & 92.17 \\
 \textbf{ALBERT large} & 56.37 & 66.20 & 99.88 & 91.53 \\
 \textbf{ALBERT xlarge} & \textcolor{red}{38.73} & 65.10 & 99.97 & 90.05 \\
 \textbf{ALBERT xxlarge} & 56.86 &  \textcolor{blue}{93.90} & 99.94 & 92.24 \\
 \hline
 \textbf{XLM-RoBERTa base} & 55.88 & 45.20 & 97.64 & 91.74 \\
 \textbf{XLM-RoBERTa large} & 62.75 & 65.45 & 99.95 & 91.74 \\
 \hline
 \textbf{XLNet base} & 58.33 & 55.00 & 99.19 &  \textcolor{blue}{94.28} \\
 \textbf{XLNet large} & 58.33 & 71.40 & 99.86 & 92.94 \\
 \hline \hline
 \textbf{Largest class} & 35.29 & 25.00 & 50.02 & 50.03\\
 \hline
 \end{tabular}
\end{table}


Results show that the LMs were able to classify hypotheses with reasonable success. Almost all models came close to solving RuleTaker and achieved an accuracy above 90\% in SimpleLogic. Results were comparatively lower for FOLIO and LogicNLI, but the LMs generally surpassed the largest class baselines by a considerable margin. A weaker performance observed in these two datasets was expected, given their greater language variability and broader logical scope; and in the case of FOLIO, its smaller size as well. Overall, the encoder-only transformer models worked relatively well as {\em soft reasoners} \cite{RuleTaker}, being able to successfully deduce theorems from premises expressed in natural language. Noteworthy exceptions were the performances of AlBERT-XL in FOLIO, DeBERTa-XL and -XXL in LogicNLI and SimpleLogic, DeBERTa-XXL in RuleTaker, and the overall lower performance of DistilBERT.

\section{Cross-Probing Fine-Tuned LMs}\label{sec:probing}



The encoder-only transformer models showed reasonable performance on the hypothesis classification task, where they were fine-tuned on the logical reasoning datasets. This, however, raises some questions: has the ability to solve this task, whatever it is, been acquired during the fine-tuning stage, or was it present from the start (i.e., from pretraining)? Most importantly, have LMs truly developed a generalized logical reasoning capability?
To examine these questions, we run a \textit{cross-probing} task: we take the LMs previously fine-tuned on our logical datasets, as well as a pretrained LM, and probe them on these same datasets. 
Given the large number of possible tests, we restricted our investigation to a single model, RoBERTa-large, so as to dig deeper on it.
This LM demonstrated a suitable balance between performance, consistency among datasets, and training time in the previous tests.

To start, we took the best fine-tuned RoBERTa-large model for each dataset and removed their classification heads, leaving just the transformer blocks, as in a pretrained model. Then, we attached a new classification head to it; i.e., the probe. As in the fine-tuning stage, we passed the formatted inputs in natural language to the LMs and tried to predict the correct label for a set of premises. However, unlike the fine-tuning stage, only the probe is updated now, while the model's body is kept frozen during the backward pass. The goal is to assess if some logical reasoning ability was learned by the LM without letting the model adapt to the task.

The same training policies from the fine-tuning step  were followed in this stage: models were trained with early stopping for up to 50 epochs (patience of 5 epochs) based on the validation loss, using two learning rates (1e-5 and 1e-6). Also, two different classifiers were tested as probes:

\begin{itemize}
    \item \textbf{1-layer} A single affine transformation is applied to the embedding of the [CLS] token. The classification head has shape (1024, labels); 1024 being the dimensionality of RoBERTa-large hidden states. We used a dropout of 0.5 before the classifier.
    
    \item \textbf{3-layer} The [CLS] embedding passes through three consecutive layers of shape (1024, 256), (256, 64), and (256, labels), respectively. We used a dropout of 0.5 in between linear layers and ReLU as the activation function.
\end{itemize}

In the tests, the two probes led to similar results. We take this as strong evidence that the knowledge used in the logical reasoning tasks, whatever it is, can be linearly recovered from the internal representations of RoBERTa-large.

\begin{table*}[t] 
 \caption{Results for the cross-probing task.  On the left, we present the best fine-tuned RoBERTa-large models for each dataset. The datasets used in the probes are listed at the top. We report only the best result for each probe. Blue cells indicate the probe of a fine-tuned model on the same dataset. In parentheses, we denote the percentage difference from the pretrained model. ``Largest class'' refers to the accuracy achieved by always selecting the class with the highest frequency in the training set. 
 }
 \label{tab:probing_datasets}
 
 \centering
 \begin{tabular}{c|c|c|c|c}
 \cline{2-5}
 & \textbf{FOLIO} & \textbf{LogicNLI} & \textbf{RuleTaker} & \textbf{SimpleLogic}\\ 
 \hline 
 \textbf{Pretrained} & 32.88 & 25.54 & 50.01 & 61.19 \\
 \hline
 \textbf{FOLIO} & \cellcolor{blue!15}55.05 (\textcolor{blue}{+67.42}) & 27.28 (\textcolor{blue}{+6.81}) & 60.74 (\textcolor{blue}{+21.45}) & 62.82 (\textcolor{blue}{+2.66})\\
 \textbf{LogicNLI} & 36.60 (\textcolor{blue}{+11.31}) & \cellcolor{blue!15}67.95 (\textcolor{blue}{+166.05}) & 69.37 (\textcolor{blue}{+38.70}) & 62.06 (\textcolor{blue}{+1.42})\\
 \textbf{RuleTaker} & 40.32 (\textcolor{blue}{+22.62}) & 36.01 (\textcolor{blue}{+40.99}) & \cellcolor{blue!15}99.44 (\textcolor{blue}{+98.84}) & 62.89 (\textcolor{blue}{+2.77})\\
 \textbf{SimpleLogic} & 32.88 (0) & 25.44 (\textcolor{red}{-0.39}) & 51.02 (\textcolor{blue}{+2.01}) & \cellcolor{blue!15} 92.35 (\textcolor{blue}{+50.92})\\
 \hline
 \textbf{Largest class} & 35.29 & 25.00 & 50.02 & 50.03\\
 \hline
 \end{tabular}
\end{table*}

Table \ref{tab:probing_datasets} displays the best results attained in the cross-probing task. The left column lists the RoBERTa-large fine-tuned models from the previous experiment, while the remaining columns represent the datasets they were probed against. The blue cells along the diagonal indicate instances where models were probed on the same datasets they were initially fine-tuned on. Percentage differences in accuracy relative to the pretrained case are reported in parentheses. As expected, the fine-tuned model for a specific dataset performed better in that same dataset, albeit less than in the fine-tuning scenario. This makes sense, since in fine-tuning, both the model's body and head are optimized, whereas in a probe the head is in charge of all the learning. These results serve as a sanity check that our probing is working.

The first row in Table \ref{tab:probing_datasets} contains the scores for the probes with pretrained RoBERTa. Accuracy levels for LogicNLI and RuleTaker closely resemble the largest class baseline, while the result for FOLIO falls below its corresponding baseline. Pretrained RoBERTa only helped to solve SimpleLogic, a dataset constrained to conjunctive implication with minimal language variation. Its pretraining scheme, dynamic masked language modeling, did not equip it with adequate logic-like knowledge to solve complex reasoning problems without specific training.

We can see from this that pretrained RoBERTa appears to have no proper logical reasoning skills. But has it acquired such an ability through fine-tuning on logical datasets? After all, RoBERTa-large was able to solve the hypothesis classification tasks reasonably well after specific training. When analyzing the results of the cross-probing task, however, we may doubt whether a general logical reasoning ability in fact emerged from fine-tuning.


In general, the fine-tuned LMs showed limited transferability when probed on different datasets. Although some gain was achieved compared to the pretrained model, they remained well below what an LM fine-tuned on the same dataset could obtain. SimpleLogic presents an interesting case. The LMs fine-tuned on the other datasets performed similarly to the pretrained model on this dataset. This is despite SimpleLogic covering only a subset of propositional calculus, a domain included in those datasets. One would expect that a model capable of solving more complex problems would be able to reason on this simpler dataset (in terms of logical scope and linguistic variability). At the same time, the LM fine-tuned on SimpleLogic did not exhibit improved performance on the other datasets, indicating a lack of acquired general logical reasoning ability during its training.

Two main conclusions can be drawn from this experiment:
\begin{enumerate}
    \item If the difference in accuracy between pretrained RoBERTa and the largest class baseline indicates the amount of logical reasoning contained in this LM, then pretrained RoBERTa seems to have very limited or no logical reasoning abilities.
    \item If the difference in accuracy between fine-tuned LMs (when applied to different datasets) and the pretrained RoBERTa model indicates the amount of logical reasoning they acquired in the fine-tuning process, then these LMs have acquired little or no general logical reasoning capability as well, suggesting that they mostly learned statistical features of the datasets. This aligns with other findings for transformer models \cite{SimpleLogic,dziri2023faith}.
\end{enumerate}




\section{Inspecting Fine-Tuned Models Layerwise}\label{section:locus}



We now address another question: which parts of the LMs are more capable of solving logical reasoning tasks? To answer this, we probe the different layers of the fine-tuned RoBERTa-large models using the same datasets they were trained on. Our goal is to identify which layers are more effective in deducing hypotheses. We expect this to provide further evidence of what sort of knowledge LMs are using to solve the hypothesis classification task. Similar to the previous experiment, the fine-tuned LMs were frozen, the older classification head was removed, and a probe was trained on top of the layers. More concretely, for each layer \textit{i} in the model, we passed the premises through the model up to layer \textit{i}, and used the outputted embedding of the [CLS] token at that layer as the input to the probe. As in the previous experiment, only the probe was trained.
The same two classifiers from the last experiment were tested. They were positioned on top of the [CLS] token of each layer, with 25 layers in total (24 transformer blocks plus the initial embedding layer). We adopted the same configurations from the previous tests: models were trained with early stopping for up to 50 epochs (with a patience of 5 epochs) based on the validation loss, using two learning rates (1e-5 and 1e-6).


\begin{figure*}[t]
 \centering
 \hspace*{-1.2cm}
 \includegraphics[width=1.2\textwidth]{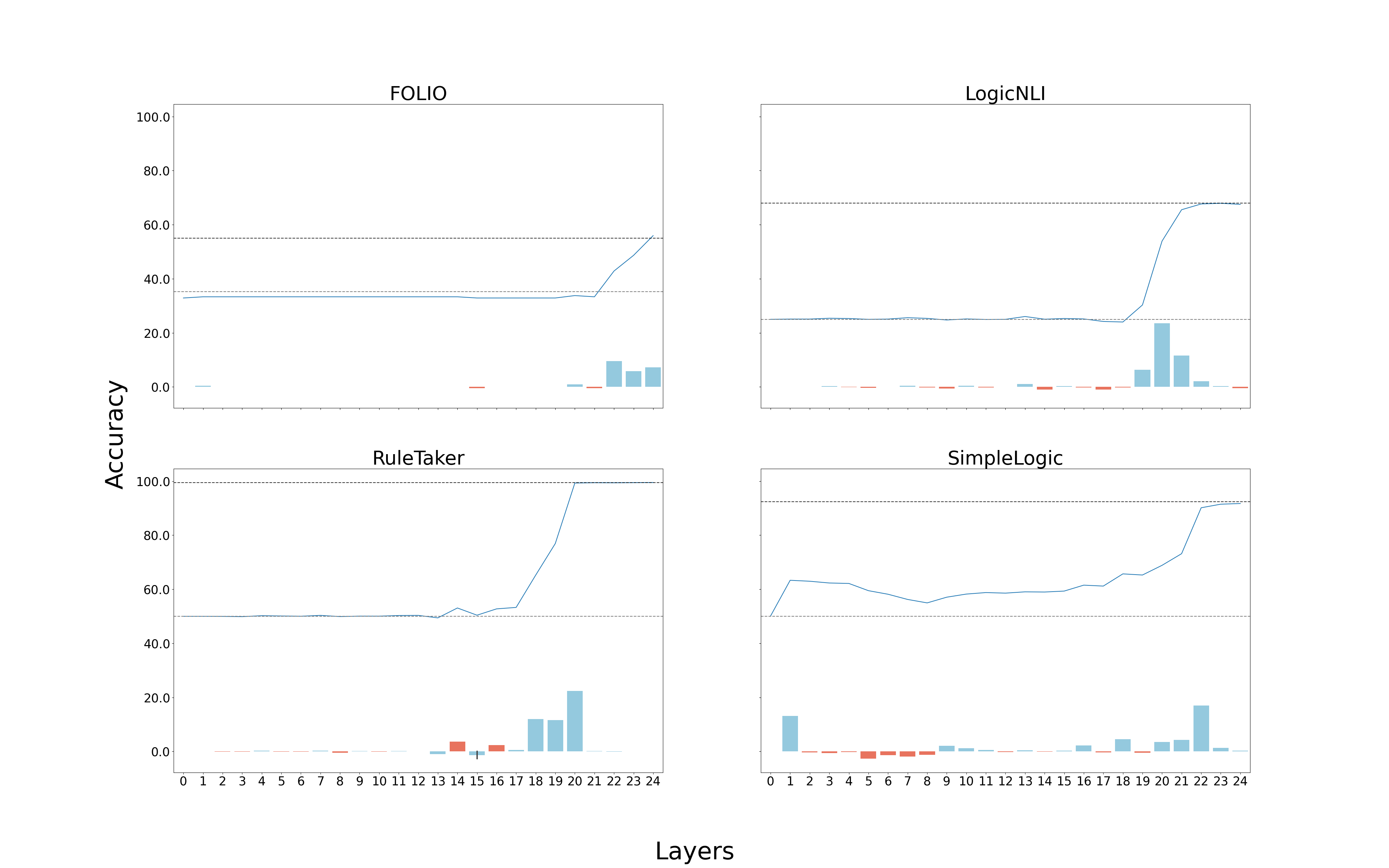}
 \caption{RoBERTa-large models fine-tuned on FOLIO, LogicNLI, RuleTaker, and SimpleLogic, and probed for the same datasets layerwise. The pretrained baselines are indicated by gray lines, while the values achieved in the cross-probing task are represented by black lines. The colored bars indicate the change in accuracy from the previous layers. Probing was performed with a 1-layer classifier and a learning rate of 1e-6.}
 \label{fig:probing_layer}
\end{figure*}

Figure \ref{fig:probing_layer} shows the accuracy for the probes stacked on the various layers of the fine-tuned LMs, using a 1-layer classifier and a learning rate of 1e-6 (Appendix \ref{appendix:layerwise} presents graphs for the other configurations). The blue line plots the accuracy on the task for each layer, while the bars display the differential score for layer$_i$; i.e., the change in accuracy from layer$_i$ to layer$_{i-1}$ \cite{bertrediscovers}. The gray line marks the pretrained model baseline, and the black line indicates the score achieved in the cross-probing task. 

A similar behavior is exhibited by all models. They remain close to the pretrained baseline in the low and mid layers. Accuracy then grows rapidly in the final layers, achieving a performance equal to the cross-probing baseline. How does this compare to other types of knowledge found in transformer models? Literature on transformers shows that surface information, such as sentence length \cite{jawahar-etal-2019-bert}, is mostly captured by lower layers. Middle layers are responsible for processing syntactic knowledge, like syntax trees \cite{hewitt-manning-2019-structural}. Finally, higher layers are responsible for task-specific functions \cite{visualizingbert2019} and contextual representations \cite{contextualrepresentations2019}.

In our layerwise probing, higher layers were the only ones able to solve the hypothesis classification task better than a pretrained model. This suggests that the knowledge acquired during fine-tuning was connected to dataset-specific features rather than general representations. It also explains why the information was not transferable among datasets. Although indirect, this experiment provides further evidence that encoder-only transformer models do not possess robust logical reasoning capabilities.
SimpleLogic was the only case that presented a growth in the initial layers. This behavior may indicate that the dataset is solvable through the use of some heuristics based on shallow statistical features, such as the number of premises, as discussed by \cite{SimpleLogic}. 

\section{Conclusion}
Logical reasoning is a valuable ability that humans use in thinking, arguing, and planning, as well as a core component of many AI systems. Here, we investigated the role of logical reasoning in encoder-only transformer models. By gathering a number of logical reasoning datasets, we observed that language models can be trained to perform complex logical tasks with relative success. However, upon closer inspection, doubts arose regarding whether these models have truly learned to reason according to logical rules. First, by probing a pretrained RoBERTa-large model with logical reasoning datasets, it became apparent that this language model did not possess intrinsic logical reasoning abilities. Second, models fine-tuned on one dataset struggled to generalize well to other datasets, even within the same domain. This observation suggests that these language models did not acquire robust logical reasoning capabilities even after specific training. Third, the knowledge necessary to solve logical reasoning tasks seems to emerge primarily at higher, more contextual layers, probably linked to statistical features of the datasets rather than deeper representations.


\section{Limitations}

We run experiments for a large variety of encoder-only transformer models in Section \ref{sec:hypothesis}. However, due to space and time constraints, we focused on RoBERTa-large for the analysis in Sections \ref{sec:probing}-\ref{section:locus}.
While we expect the same behavior to appear in the other encode-only models, further tests are needed to verify whether conclusions can be reliably extended to them. We have not explored decoder nor encoder-decoder models either, which could widely extend the number of models to be tested. We cannot rule out the possibility that robust logical reasoning is an emergent ability only manifested in large language models \cite{emergent2022}. Additionally, other types of representations, such as attentions and feedforward computations, could be analyzed in relation to logical reasoning. Further work should also focus on other types of logical formalism beyond PC and FOL.

%
\section*{Acknowledgements}

This work was supported by the Center for Artificial Intelligence USP/IBM/FAPESP (C4AI), 
jointly funded by the S\~ao Paulo Research Foundation (FAPESP grant 2019/07665-4) and by 
the IBM Corporation. Research by Marcos José has been carried out with support by \textit{Ita\'{u} Unibanco S.A.} through the scholarship program  \textit{Programa de Bolsas Ita\'{u}} (PBI) ;
Fabio Cozman was partially supported by CNPq grants 312180/2018-7 and 305753/2022-3. 
We acknowledge support also by CAPES - Finance Code 001.

\bibliographystyle{splncs04}
\bibliography{custom}

\clearpage

\appendix

\section{Datasets}\label{appendix:datasets}

In this appendix, we describe the four logical reasoning datasets in more detail. Table \ref{tab:data_source} indicates the sources from where they were obtained.

\paragraph{FOLIO} \cite{FOLIO} is a human-annotated dataset for FOL reasoning problems. Logically-sound contexts were generated in two ways: in the first, annotators created contexts from scratch, based on random Wikipedia pages; in the second, a template of nested syllogisms was used, from which annotators then replaced the abstract entities and categories by nouns, phrases or clauses, as to make the text to reflect real-life scenarios. Next, the authors verified the alignment between natural language sentences and FOL formulas and added implicit commonsense knowledge as premises. After that, they verified the syntactic validity and label consistency of FOL formula annotations with a FOL prover. Finally, sentences were reviewed for grammar issues and language fluency. Only train and validation tests are available, so we used the latter for reporting tests. Hypotheses can be \textit{True}, \textit{False}, or \textit{Unknown}.

\paragraph{LogicNLI} \cite{LogicNLI} is a FOL dataset created through a semi-automatic method. A set of logical templates was defined and then filled by subjects and predicates sampled from predefined sets. Next, manual edits were made to correct grammatical errors and resolve semantic ambiguities. Hypothesis are classified as \textit{Entailment}, \textit{Contradiction}, \textit{Neutral}, and \textit{Paradox}. A paradox is defined as a situation where both a sentence and its contrary can be inferred from the premises. We used the standard version of the dataset, which encompasses all labels.

\paragraph{RuleTaker} \cite{RuleTaker} is a logical reasoning dataset in which rules are conjunctive implications. Predicates may be negated and facts may be either attributes (which assign properties to entities) or relations (which relate two entities). We used the ParaRules version, where rules and facts were paraphrased by crowdworkers into more natural language; paraphrased constructions were then combined to form new templates. We also used the updated version of RuleTaker (problog), which eliminated some world model inconsistencies.
Hypothesis can be \textit{True}, when a hypothesis follows from the premises, and \textit{False} otherwise (closed-world assumption, CWA).

\paragraph{SimpleLogic} \cite{SimpleLogic} is similar to RuleTaker, only supporting conjunctive implication (facts are simply conjunctive implications with empty antecedents). Language variance is virtually removed by using a fixed template for translating FOL into natural language and by the use of a small random list of words as predicates. Argumentative complexity is limited by setting thresholds for input length, number of predicates, and reasoning depth. We reconstructed the original template based on the examples given in the paper. For our tests, we used the RP Balanced version. We undersampled the largest class (True) to obtain the same number of observations as for the False class. Labels can be \textit{True} and \textit{False} (CWA).

\begin{table}[H] \small
    \centering
    \begin{tabular}{l|l} 
 \textbf{Dataset} & \textbf{Source}\\
 \hline
 FOLIO & \url{https://github.com/Yale-LILY/FOLIO}\\
 LogicNLI & \url{https://github.com/omnilabNL/LogicNLI}\\
 RuleTaker & \url{https://allenai.org/data/ruletaker}\\
 SimpleLogic & \url{https://github.com/joshuacnf/paradox-learning2reason}\\
 \hline
    \end{tabular}
    \caption{Sources for the datasets.}
    \label{tab:data_source}
\end{table}

\section{Label distribution}\label{appendix:label}
Table \ref{tab:label_distribution} shows the label distribution for the datasets used in the paper. The row above provides the number of observations per label, and the row below shows their relative percentage. Labels: \textbf{FOLIO}: False, True, Unknown. \textbf{LogicNLI}: Contradiction, Entailment, Neutral, Paradox. \textbf{RuleTaker}: False, True. \textbf{SimpleLogic}: False, True.

\begin{table}[H] \small
 \centering
 \begin{tabular}{l|l|l|l}
 & \textbf{Train} & \textbf{Validation} & \textbf{Test} \\
 \hline
 FOLIO & 286 / 388 / 329 & - & 63 / 72 / 69 \\ & (28.51\% / 38.68\% / 32.80\%) & - & (30.88\% / 35.29\% / 33.82\%) \\ 
 LogicNLI & 4000 / 4000 / 4000 / 4000 & 500 / 500 / 500 / 500 & 500 / 500 / 500 / 500 \\ & (25\% each) & (25\% each) & (25\% each) \\ RuleTaker & 13666 / 13697 & 1946 / 1953 & 3895 / 3898 \\ & (49.94\% / 50.05\%) & (49.91\% / 50.09\%) & (49.98\% / 50.02\%) \\ 
 SimpleLogic & 5696 / 5645 & 683 / 735 & 709 / 708 \\ & (50.22\% / 49.77\%) & (48.16\% / 51.83\%) & (50.03\% / 49.96\%)\\
\hline
 \end{tabular}
  \caption{Label distribution for logical reasoning datasets.\label{tab:label_distribution}}
\end{table}

\section{Laywerwise probing}\label{appendix:layerwise}
\begin{figure}[H]
 \hspace*{-1.2cm}
 \includegraphics[width=1.2\textwidth]{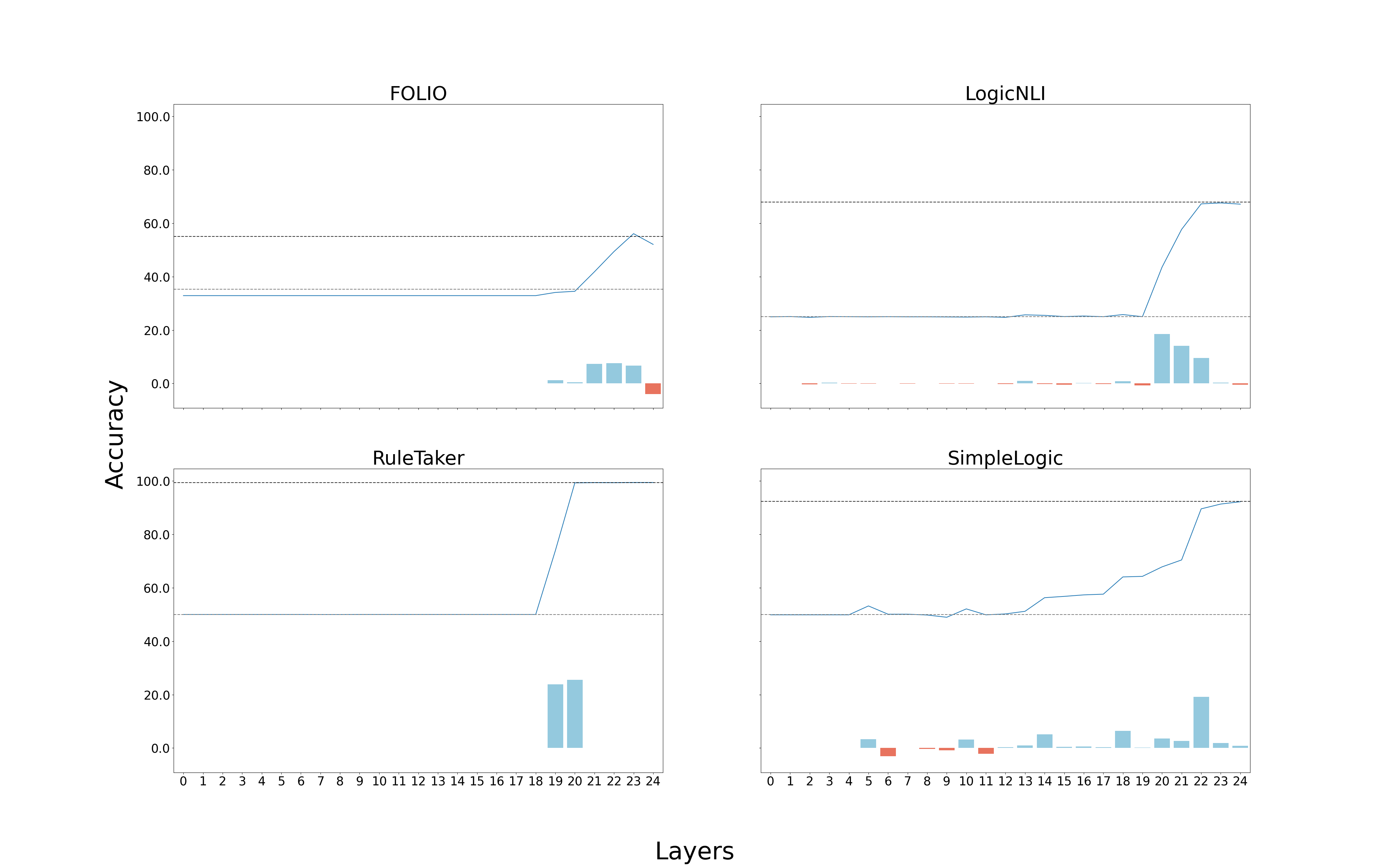}
 \caption{RoBERTa-large models fine-tuned on FOLIO, LogicNLI, RuleTaker, and SimpleLogic, and probed for the same datasets layerwise. The pretrained baselines are indicated by gray lines, while the values achieved in the cross-probing task are represented by black lines. The colored bars indicate the change in accuracy from the previous layers. Probing was performed with a 3-layer classifier and a learning rate of 1e-6.}
 \label{fig:probing_layer_3layer_06}
\end{figure}

\begin{figure}[H]
 \hspace*{-1.2cm}
 \includegraphics[width=1.2\textwidth]{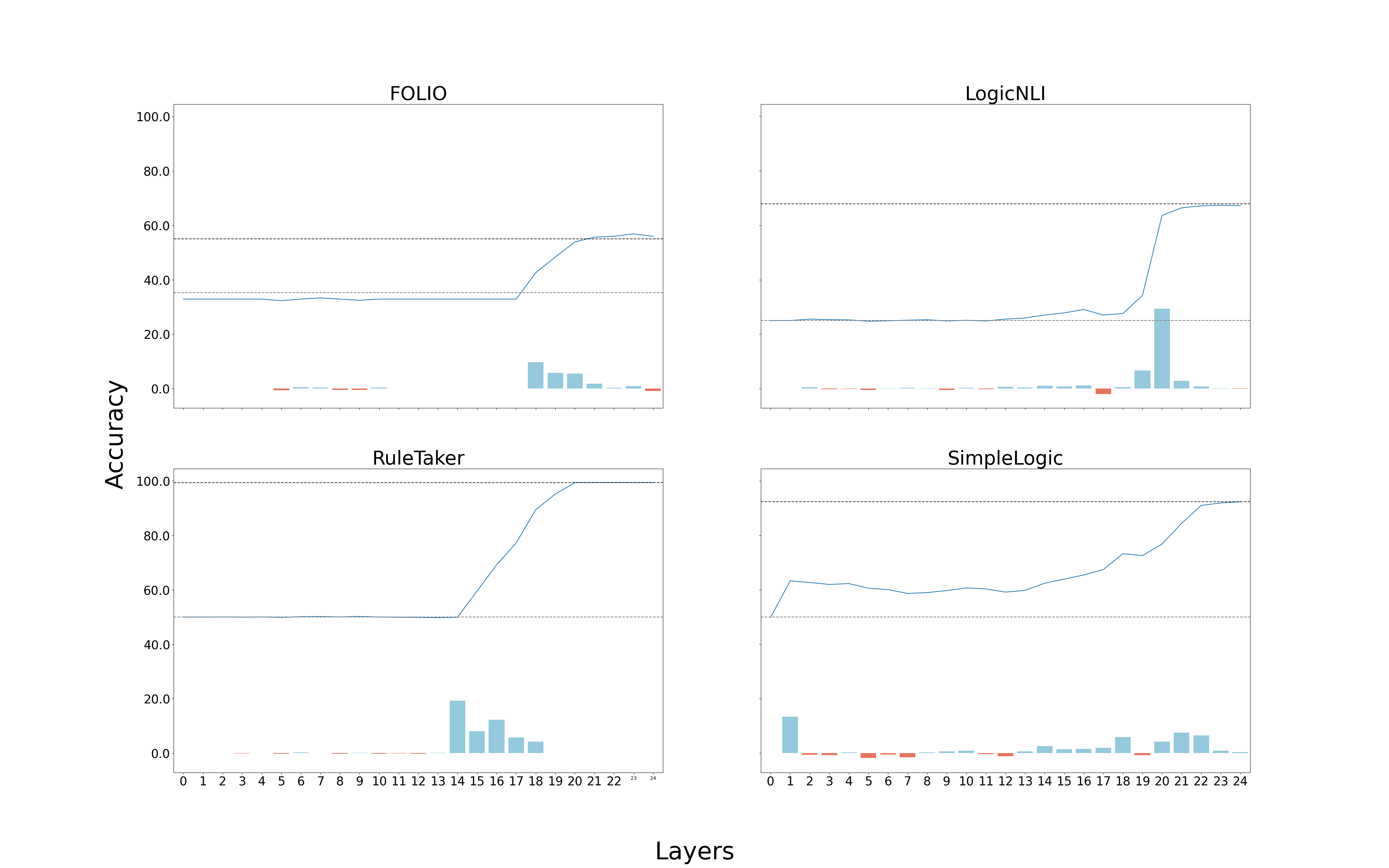}
 \caption{RoBERTa-large models fine-tuned on FOLIO, LogicNLI, RuleTaker, and SimpleLogic, and probed for the same datasets layerwise. The pretrained baselines are indicated by gray lines, while the values achieved in the cross-probing task are represented by black lines. The colored bars indicate the change in accuracy from the previous layers. Probing was performed with a 1-layer classifier and a learning rate of 1e-5.}
 \label{fig:probing_layer_1layer_05}
\end{figure}

\begin{figure}[H]
 \hspace*{-1.2cm}
 \includegraphics[width=1.2\textwidth]{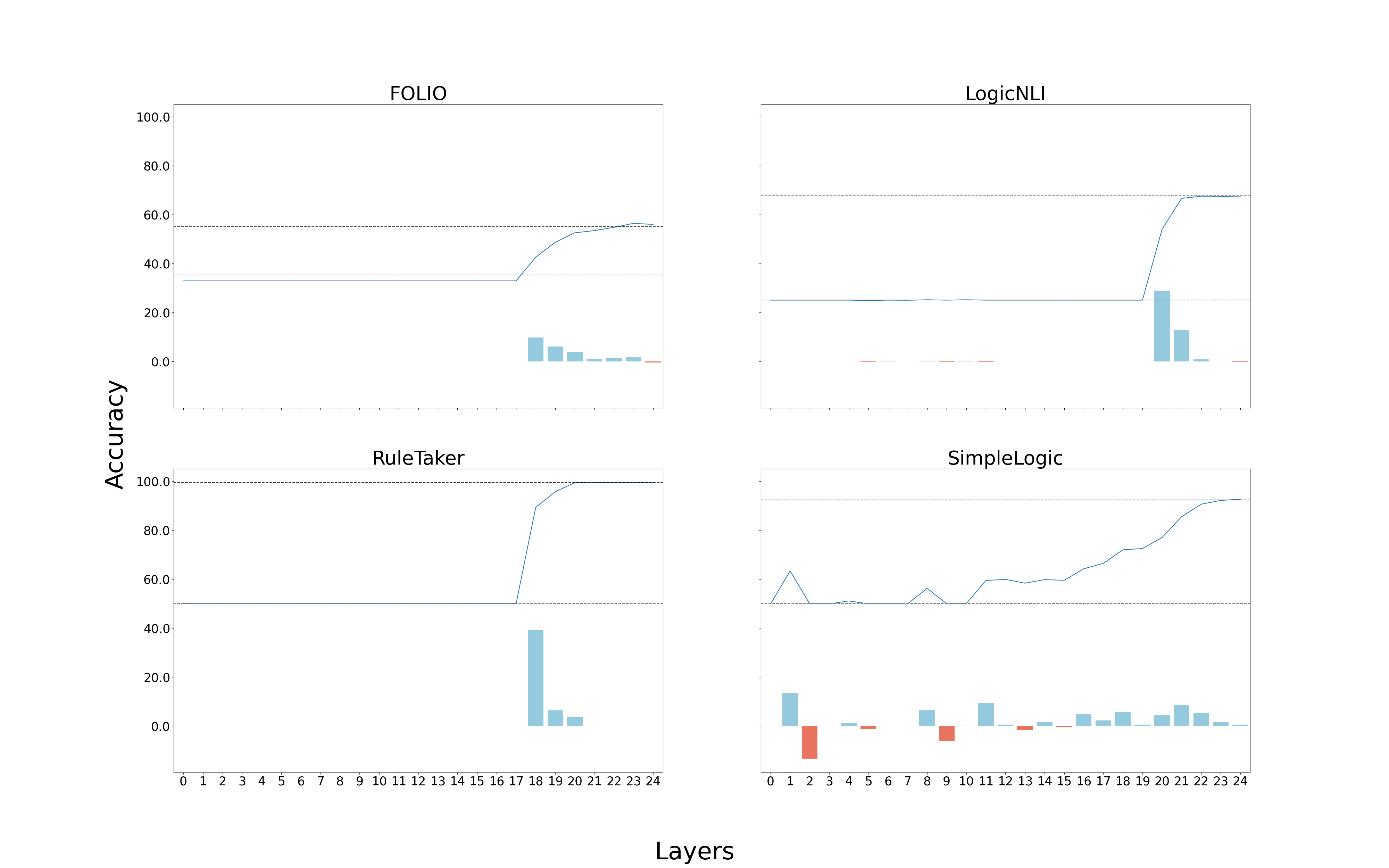}
 \caption{RoBERTa-large models fine-tuned on FOLIO, LogicNLI, RuleTaker, and SimpleLogic, and probed for the same datasets layerwise. The pretrained baselines are indicated by gray lines, while the values achieved in the cross-probing task are represented by black lines. The colored bars indicate the change in accuracy from the previous layers. Probing was performed with a 3-layer classifier and a learning rate of 1e-5.}
 \label{fig:probing_layer_3layer_05}
\end{figure}

For the layerwise probing (Sec. \ref{section:locus}), we tested two different classifiers (1-layer and 3-layer) and two learning rates (1e-6 and 1e-5). Figure \ref{fig:probing_layer} above displayed the results for the 1-linear classifier and 1e-6 learning rate. The figures below display the results for the other probes. Figure \ref{fig:probing_layer_3layer_06} provides the graphs for the 3-layer classifier and learning rate of 1e-6; Figure \ref{fig:probing_layer_1layer_05} provides the graphs for the 1-layer classifier and learning rate of 1e-5; and Figure \ref{fig:probing_layer_3layer_05} provides the graphs for the 3-layer classifier and learning rate of 1e-5.

\clearpage

\end{document}